\documentclass[preprint]{vgtc}               

\ifpdf
  \pdfoutput=1\relax                   
  \usepackage{graphicx}                
  \DeclareGraphicsExtensions{.pdf,.png,.jpg,.jpeg} 
\else
  \usepackage{graphicx}                
  \DeclareGraphicsExtensions{.eps}     
\fi%

\graphicspath{{figures/}{pictures/}{images/}{./}} 

\usepackage{microtype}                 
\PassOptionsToPackage{warn}{textcomp}  
\usepackage{textcomp}                  
\usepackage{times}                     
\usepackage{cite}                      
\usepackage{tabu}                      
\usepackage{booktabs}                  

\preprinttext{To appear in IEEE VR 2022}

\onlineid{xxxx}

\vgtccategory{Research}

\vgtcinsertpkg

\usepackage{float}
\usepackage{longtable,booktabs}
\usepackage[ruled,vlined]{algorithm2e}
\usepackage{multirow}
\usepackage{amsmath}
\usepackage{amsfonts}
\usepackage{mathtools}
\usepackage{xcolor} 

\DeclareMathOperator*{\argmin}{argmin}
\DeclareMathOperator*{\argmax}{argmax}
\newcommand{\norm}[1]{\left\lVert#1\right\rVert} 
\renewcommand{\vec}[1]{\boldsymbol{#1}}

\newcommand{\x}{\vec{x}}

\newcommand{\lp}{L_\textsubscript{p}}

\def\eg{\emph{e.g}.} 
\def\ie{\emph{i.e}.}

\def\etal{\emph{et al}.}

\makeatletter
\newcommand{\thickhline}{%
	\noalign {\ifnum 0=`}\fi \hrule height 1pt
	\futurelet \reserved@a \@xhline
}
\makeatother

\title{SPAA: Stealthy Projector-based Adversarial Attacks on Deep Image Classifiers}

\author{Bingyao Huang\thanks{College of Computer and Information Science, Southwest University, Chongqing, China. E-mail: bhuang@swu.edu.cn}
\and Haibin Ling\thanks{Department of Computer Science, Stony Brook University, Stony Brook, NY 11794, USA. E-mail: hling@cs.stonybrook.edu}
}

\abstract{
	Light-based adversarial attacks use spatial augmented reality (SAR) techniques to fool image classifiers by altering the physical light condition with a controllable light source, \eg, a projector. Compared with physical attacks that place hand-crafted adversarial objects, projector-based ones obviate modifying the physical entities, and can be performed transiently and dynamically by altering the projection pattern. However, subtle light perturbations are insufficient to fool image classifiers, due to the complex environment and project-and-capture process. Thus, existing approaches focus on projecting clearly perceptible adversarial patterns, while the more interesting yet challenging goal, stealthy projector-based attack, remains open. In this paper, for the first time, we formulate this problem as an end-to-end differentiable process and propose a Stealthy Projector-based Adversarial Attack (SPAA) solution. In SPAA, we approximate the real Project-and-Capture process using a deep neural network named PCNet, then we include PCNet in the optimization of projector-based attacks such that the generated adversarial projection is physically plausible. Finally, to generate both robust and stealthy adversarial projections, we propose an algorithm that uses minimum perturbation and adversarial confidence thresholds to alternate between the adversarial loss and stealthiness loss optimization. Our experimental evaluations show that SPAA clearly outperforms other methods by achieving higher attack success rates and meanwhile being stealthier, for both targeted and untargeted attacks.
}

\CCScatlist{
  \CCScatTwelve{Human-centered computing}{Human computer interaction (HCI)}{Interaction paradigms}{Mixed / augmented reality}; 
  \CCScatTwelve{Security and privacy}{Human and societal aspects of security and privacy}{Privacy protections}{}
  \CCScatTwelve{Computing methodologies}{Artificial intelligence}{Computer vision}{Object recognition}
}

\teaser{
  \centering
  \includegraphics[width=1\linewidth]{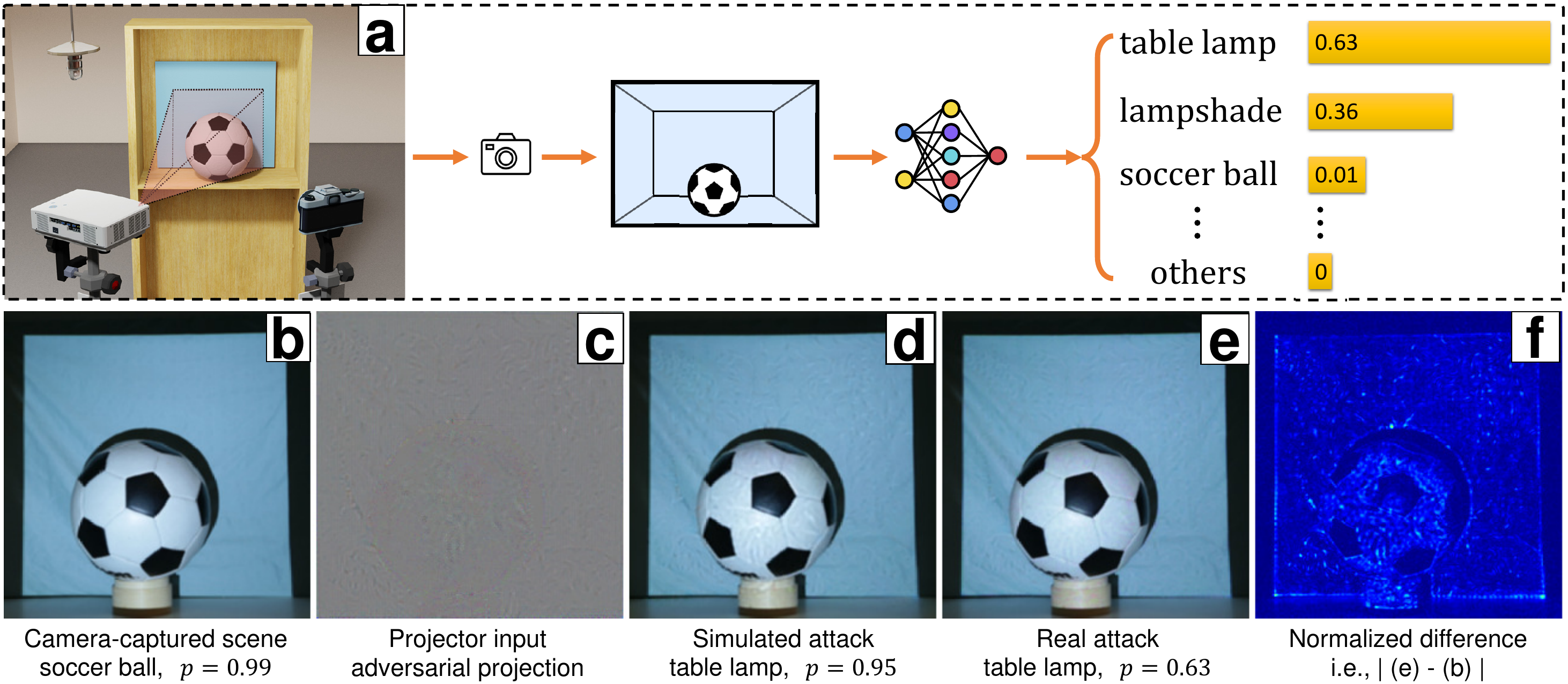}
  \caption{Stealthy projector-based adversarial attack (SPAA): \textbf{(a)} System setup: the goal is to project a stealthy adversarial pattern (\eg, (c)), such that the camera-captured scene (\eg, (e)) causes misclassification.
  \textbf{(b)} Camera-captured scene under normal light and the classifier output is \textbf{soccer ball} with a probability of $ p=0.99 $. \textbf{(c)} An adversarial pattern created by our SPAA algorithm. \textbf{(d)} Our SPAA \emph{simulated} camera-captured adversarial projection (\ie, (c) virtually projected onto (b)). \textbf{(e)} The \emph{actual} camera-captured adversarial projection (\ie, (c) actually projected onto (b)). \textbf{(f)} Normalized difference between (b) and (e). It is clear that the camera-captured adversarial projection is stealthy, meanwhile, successfully fools the classifier such that the output is \textbf{table lamp} with a probability of $ p=0.63 $. More results are provided in \autoref{sec:experiments} and supplementary.
  }\label{fig:teaser}
}

\begin{document}


\firstsection{Introduction}\label{sec:intro}

\maketitle

Adversarial attacks on deep image classifiers aim to generate adversarial perturbation to the input image (\ie, digital attacks) or the physical world (physical or projector-based attacks) such that the perturbed input can fool classifiers. With the rapid advancement of artificial intelligence, adversarial attacks become particularly important as they may be applied to protect user privacy and security from unauthorized visual recognition. It is worth noting that our work is different from existing studies in privacy and security of virtual reality (VR) and augmented reality (AR) \cite{david2021privacy, al2021vr, george2019investigating,roesner2014security, sekhavat2016privacy}, because we aim to use spatial augmented reality (SAR) to protect privacy and security rather than studying the privacy and security of VR/AR systems themselves.
The most popular type of adversarial attacks are digital attacks~\cite{szegedy2014intriguing, goodfellow2015explaining, moosavi2016deepfool, moosavi2017universal, carlini2017towards, dong2018boosting, madry2018towards, su2019one, rony2019decoupling, zhao2020towards}, which directly perturb the input images of a classifier. A common requirement for digital attack is stealthiness, \ie, the perturbation should be relatively small (usually bounded by $ \lp $ norm) yet still successfully fools the classifiers. 
Another type is physical attack~\cite{sharif2016accessorize, kurakin2017adversarialw, kurakin2017adversarial, brown2017adversarial, athalye2018synthesizing, eykholt2018robust, zeng2019adversarial, wu2020making, duan2020adversarial}, which assumes no direct access to the classifier input image. Instead, the perturbation is made on the physical entities, \eg, placing adversarial patches, stickers or 3D printed objects. Usually physical attacks are much harder to achieve stealthiness due to complex physical environment and image capture process~\cite{kurakin2017adversarialw, athalye2018synthesizing, zeng2019adversarial}, and they must be strong enough to fool the classifiers. Another challenge is for targeted attacks, physical ones must manufacture a different adversarial pattern for each target.

Light-based (in the rest of the paper, we use \textit{projector-based} to better describe our setup) attacks, as shown by our example in \autoref{fig:teaser}, use SAR techniques to modify the environment light without physically placing adversarial entities to the scene. Thus, the attacks can be transient and dynamic, \eg, by turning on and off the projector or changing the projected patterns. However, similar to physical attacks, projector-based attacks are difficult to fool image classifiers due to the complex environment and the project-and-capture process. Thus, existing methods \cite{nichols2018ProjectingTL, nguyen2020adversarial, li2020light} focus on improving attack success rates using \textit{perceptible} patterns, while \textit{stealthy} projector-based attack remains an open problem. 

Note that simply projecting a digital adversarial example to the scene may not produce a successful stealthy projector-based attack, due to the complex geometric and photometric transformations involved in the project-and-capture process. 
One intuitive solution is to use a two-step pipeline by first performing digital attacks on the camera-captured scene image, then using projector compensation techniques~\cite{huang2019compennet++, grundhofer2015robust, bimber2008visual} to find the corresponding projector adversarial pattern. However, this two-step method is problematic, because digital attacks may generate physically implausible~\cite{zeng2019adversarial} adversarial examples that cannot be produced by a projector, \eg, perturbations in shadow regions or luminance beyond the projector's dynamic range. As will be shown in our experimental evaluations, such a two-step method has lower attack success rates and stealthiness than our SPAA solution. 
Another idea is the online one-pixel-based attack~\cite{nichols2018ProjectingTL}. However, this preliminary exploration only allows to perturb one projector pixel and requires at least hundreds of real projections and captures to attack a single $32\times32$ low resolution target, making it hardly applicable to higher resolution images in practice, as shown in our experiments.

In this paper, we approach stealthy projector-based attacks from a different perspective by approximating the real Project-and-Capture process using a deep neural network named \textit{PCNet}. Then, we concatenate PCNet with a deep image classifier such that the entire system is end-to-end differentiable. Thus, PCNet adds additional constraints such that the projected adversarial patterns are physically plausible. Finally, to generate robust and stealthy adversarial patterns, we propose an optimization algorithm that uses minimum perturbation and adversarial confidence thresholds to alternate between the minimization of adversarial loss and stealthiness loss. 

To validate the effectiveness of the proposed SPAA algorithm, we conduct thorough experimental evaluations on 13 different projector-based attack setups with various objects, for \textit{both targeted and untargeted} attacks. In all the comparisons, SPAA significantly outperforms other baselines by achieving higher success rates and meanwhile being stealthier.

Our contributions can be summarized as follows:
\begin{itemize}
	\vspace{-1.5mm}\item For the first time, we formulate the stealthy projector-based adversarial attack as an end-to-end differentiable process.
	\vspace{-1.5mm}\item Based on our novel formulation, we propose a deep neural network named PCNet to approximate the real project-and-capture process.
	\vspace{-1.5mm}\item By incorporating the novel PCNet in projector-based adversarial attacks, our method generates physically plausible and stealthy adversarial projections.
\end{itemize}
The source code, dataset and experimental results are made publicly available at \url{https://github.com/BingyaoHuang/SPAA}.

In the rest of the paper, we introduce the related work in \autoref{sec:related_work}, and describe the problem formulation and the proposed SPAA algorithm in \autoref{sec:methods}. We show our system configurations and experimental evaluations in \autoref{sec:experiments}, and conclude the paper in \autoref{sec:conclusion}.

\section{Related Work}\label{sec:related_work}
\begin{figure}[!t]
	\begin{center}
		\includegraphics[width=1\linewidth]{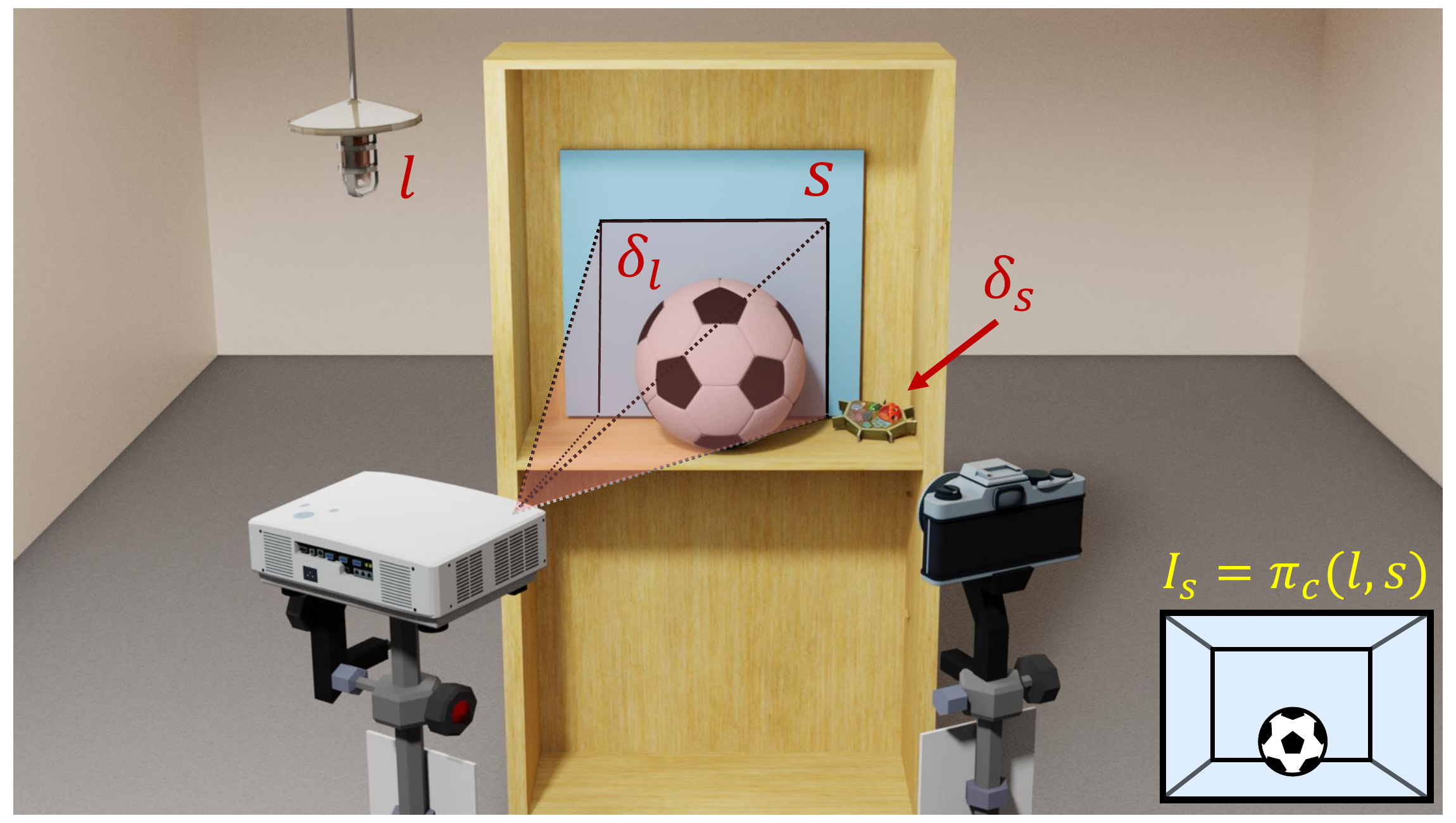}
		\caption{Adversarial attack types. Digital attacks directly perturb the camera-captured image $I_s$. Physical attacks perturb the scene $s$ by adding physical entities, \eg, an adversarial patch $\delta_s$. Projector-based attacks perturb the environment light $l$ by $\delta_l$.}\label{fig:attack_types}
	\end{center}
\end{figure}

In this section we review existing adversarial attacks on deep image classifiers in three categories: digital attacks, physical ones and projector-based ones as shown in~\autoref{fig:attack_types}.

\noindent\textbf{Digital attacks}
directly alter a classifier's input digital image such that the classifier's prediction becomes either (a) a specific target (targeted attack) or (b) any target as long as it is not the true label (untargeted attack). The input image perturbation is usually performed by back-propagating the gradient of adversarial loss to the input image, and can be either single-step, \eg, fast gradient sign method (FGSM) \cite{goodfellow2015explaining}, or iterative, \eg, L-BFGS based \cite{szegedy2014intriguing}, iterative FGSM (I-FGSM) \cite{kurakin2017adversarial}, momentum iterative FGSM (MI-FGSM) \cite{dong2018boosting}, projected gradient descent (PGD) \cite{madry2018towards}, C\&W \cite{carlini2017towards} and decoupling direction and norm (DDN) \cite{rony2019decoupling}.

The gradient-based methods above require access to the classifier weights and gradients (\ie, white-box attack). To relax such requirements, another type of digital attacks use gradient-free optimization, \eg, one-pixel attack using differential evolution (DE) \cite{su2019one} or black-box optimization \cite{zhao2019design}. 
Another advantage of gradient-free attacks is that they can be applied to scenarios where the system gradient is inaccessible or hard to compute (see projector-based attacks below). However, they are usually less efficient than gradient-based methods, and this situation deteriorates when image resolution increases.

\noindent\textbf{Physical attacks}
assume no direct access to the classifier input image, instead they modify the physical entities in the environment by placing manufactured adversarial objects or attaching stickers/graffiti. 
For example, Brown \etal \cite{brown2017adversarial} print 2D adversarial patches such that when placed in real scenes, the camera-captured images may be misclassified as certain targets. Sharif \etal \cite{sharif2016accessorize} create a pair of adversarial eyeglass frames such that wearers can evade unauthorized face recognition systems. Similarly, Wu \etal \cite{wu2020making} create an invisibility cloak to evade object detectors. Li \etal \cite{li2019adversarial} alter camera-captured scenes by applying a translucent adversarial sticker to the camera lens. Early approaches often perform attacks in the digital image space first, and then bring the printed versions to the physical world. However, Kurakin \etal \cite{kurakin2017adversarialw} show that the complex physical environment and the image capture process significantly degrade the attack success rates, because image space perturbations may not be physically meaningful \cite{zeng2019adversarial} and are sensitive to minor transformations \cite{athalye2018synthesizing}. 

To fill the gap between the digital and the physical worlds, and to improve transferability, some studies focus on robustness of physical adversarial examples against transformations. For example, Athalye \etal \cite{athalye2018synthesizing} propose Expectation Over Transformation (EOT) to generate robust physical adversarial examples over synthetic transformations. Then, Eykholt \etal \cite{eykholt2018robust} propose Robust Physical Perturbations (RP$_2$) to produce robust adversarial examples under both physical and synthetic transformations.
Afterwards, Jan \etal \cite{jan2019connecting} present D2P to capture more complex digital-to-physical transformations using an image-to-image translation network.

Despite these efforts, how to make adversarial patterns stealthy remains challenging. Unlike digital attacks where perturbations can be easily made  stealthy, subtle physical perturbations are hard to capture using digital cameras and can be easily polluted by sensor noise, lens distortion and camera internal image processing pipeline. Thus, to improve robustness against these factors, most existing physical adversarial examples are designed with strong artificial patterns.

\noindent\textbf{Projector-based attacks}
modify only the environment light condition using a projector instead of changing the physical entities (\eg, placing manufactured adversarial objects in the scene), and very few studies have been dedicated to this direction. A preliminary exploration done by Nichols and Jasper \cite{nichols2018ProjectingTL} uses a low resolution projector-camera pair (both set to $32\times32$) to perturb scene illuminations and capture projections. 
Because the image resolutions are relatively small, a differential evolution~\cite{storn1997differential} (DE)-based one-pixel attack framework \cite{su2019one} can be applied to solve this problem. In particular, by perturbing only one projector pixel, only five variables need to be optimized, \ie, the pixel's 2D location and its RGB value. Even so, it still requires hundreds of real projections and captures for each targeted attack. Moreover, including the real project-and-capture process  in the DE optimization may not only cause efficiency bottlenecks but also makes it hard to run in parallel. Thus, this method is impractical for high resolution cases due to the exponentially increased number of real project-and-capture processes. 
Other studies focus on attacking face recognition systems~\cite{zhou2018invisible, nguyen2020adversarial, shen2019vla, li2020light}. Special hardware settings are proposed to achieve stealthiness, \eg, Zhou \etal \cite{zhou2018invisible} use infrared LEDs to project human imperceptible patterns and Shen \etal \cite{shen2019vla} leverage persistence of vision and the chromatic addition rule to control camera shutter speed, such that the camera can capture human imperceptible adversarial patterns.

\noindent\textbf{Stealthiness}
is a common requirement for adversarial attacks, \ie, perturbations should be (nearly) imperceptible to human eyes while still successfully causing misclassification. Usually stealthiness is measured using $\lp$ norm \cite{szegedy2014intriguing, goodfellow2015explaining, carlini2017towards, kurakin2017adversarialw, moosavi2016deepfool} and used as an additional constraint when optimizing the adversarial attack objective. Recently, Zhao \etal \cite{zhao2020towards} show that optimizing perceptual color distance $ \Delta E $ (\ie, CIEDE2000 \cite{luo2001development}) instead of $\lp$ norm may lead to more robust attacks yet still being stealthy. Besides pixel-level color losses, neural style similarity constraints can also improve stealthiness, \eg, Duan \etal \cite{duan2020adversarial} propose an adversarial camouflage algorithm named AdvCam to make physical adversarial patterns look natural. Although it looks less artificial than previous work \cite{eykholt2018robust,brown2017adversarial}, there is still room for improvement, especially the texture and color. 

\noindent\textbf{The proposed SPAA}
belongs to projector-based attacks, and is most related to the preliminary exploration in~\cite{nichols2018ProjectingTL}, with the following main differences: (1) We formulate projector-based adversarial attack as an end-to-end differentiable process, and simulate the real project-and-capture process with a deep neural network. 
(2) With such a formulation and implementation, our method can perform projector-based attacks using gradient descent, which is more efficient than one-pixel differential evolution \cite{nichols2018ProjectingTL}. (3) Because the real project-and-capture process is excluded from the gradient descent optimization, our method is more efficient and parallelizable, and multi-classifier and multi-targeted adversarial attacks can be performed simultaneously in batch mode. (4) Our SPAA achieves much higher attack success rates, yet remains stealthy.

\section{Methods}\label{sec:methods}
\subsection{Problem formulation}
Denote $ f $ as an image classifier that maps a camera-captured image $ I $ to a vector of class probabilities $f(I) \in [0, 1]^N$, for $ N $ classes, and denote $f_i(I)\in [0, 1]$ as the probability of the $i$-th class. Typically, \textit{targeted} digital adversarial attacks aim to perturb $I$ by a small disturbance $ \delta $ whose magnitude is bounded by a small number $\epsilon>0$, such that a certain target $ t $ (other than the true label $ t_\text{true} $) has the highest probability. Similarly, \textit{untargeted} attacks are successful as long as the classifier's output label is not the true class $t_\text{true}$:
\begin{align}\label{eq:digital_tar}
	&   \argmax_i f_i(I+\delta)\begin{cases*}
			= t \quad\quad &\text{targeted}\\
			\neq t_\text{true} \quad\quad & \text{untargeted}
		\end{cases*}\nonumber \\
	& \text{subject to} \quad \mathcal{D}(I, I+\delta) < \epsilon, 
\end{align}
where $\mathcal{D}$ is a distance metric measuring the similarity between two images, \eg, $\lp $ norm, which also measures the perturbation stealthiness.

We extend \autoref{eq:digital_tar} to physical world (\autoref{fig:attack_types}) and denote the camera capture function as $ \pi_c $, which maps the physical scene $ s $ (\ie, including all geometries and materials in the scene) and lighting $ l $ to a camera-captured image $ I $ by:
\begin{equation}
	I = \pi_c(l, s)
\end{equation}
Physical adversarial attacks aim to perturb the physical entities $ s $ such that the classifier misclassifies the camera-captured image $I$ as a certain target label $t$ (or any label other than $t_\text{true}$ for untargeted attacks). 
By contrast, projector-based attacks aim to perturb the lighting $ l $ by $ \delta_l $ such that the camera-captured image causes misclassification, \ie:
\begin{align}
	&   \argmax_{i} f_i(\pi_c(l+\delta_l, s))\begin{cases*}
		 = t, \quad\quad &\text{targeted}\\
		\neq t_\text{true} \quad\quad & \text{untargeted}
		\end{cases*} \nonumber\\
	& \text{subject to} \quad \mathcal{D}\left(\pi_c(l+\delta_l, s), \pi_c(l, s)\right) < \epsilon
\end{align}
In this paper, $ \delta_l $ is illumination perturbation from a projector. Denote the projector's projection function and input image as $ \pi_p $ and $x$, respectively. Then, the illumination generated by the projector is given by $ \delta_l = \pi_p(x) $, and the camera-captured scene under superimposed projection is given by $I_x = \pi_c(l + \pi_p(x), s)$. 
Denote the composite project-and-capture process above (\ie, $\pi_c$ and $\pi_p$) as $ \pi:x \mapsto I_x $, then the camera-captured scene under superimposed projection is:
\begin{equation}
	I_x = \pi(x, l, s)
\end{equation}
Finally, projector-based adversarial attack is to find a projector input adversarial image $x'$ such that:
\begin{align}\label{eq:real_pc_obj}
	&   \argmax_i f_i\left(I_{x'} = \pi(x', l, s)\right)
		\begin{cases*}
		 = t, \quad\quad &\text{targeted}\\
		 \neq t_\text{true} \quad\quad & \text{untargeted}
		\end{cases*} \nonumber\\
	& \text{subject to} \quad \mathcal{D}\left(I_{x'}, I_{x_0}\right) < \epsilon,
\end{align}
where $ x_0 $ is a null projector input image. 

\begin{figure*}[!t]
	\begin{center}
		\includegraphics[width=1\linewidth]{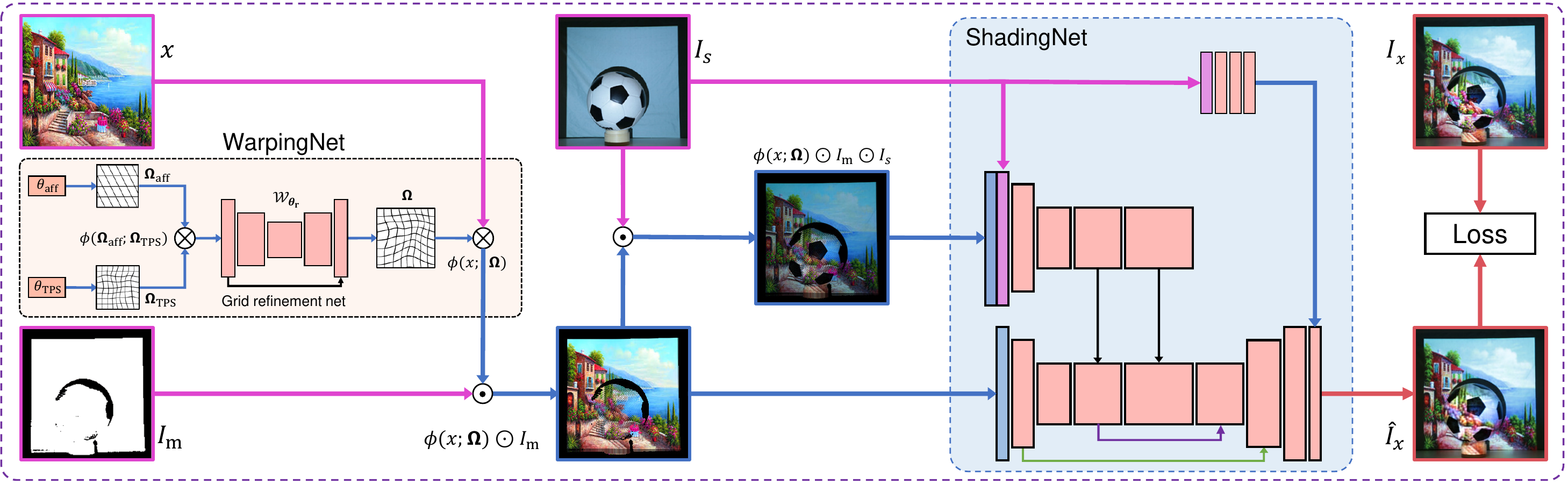}
		\caption{PCNet $ \hat{\pi} $ architecture and training. PCNet approximates the real project-and-capture process $\pi$ using a deep neural network (WarpingNet + ShadingNet). The inputs are a projector input image $x$, a camera-captured scene image (under normal light) $I_s$, and a projector direct light mask $I_\text{m}$. The output $\hat{I}_x$ is an inferred camera-captured scene (under superimposed projection). \textbf{WarpingNet} consists of a learnable affine matrix $ \theta_\text{aff} $, thin-plate-spline (TPS) parameters $ \theta_\text{TPS} $ and a grid refinement network $\mathcal{W}_{\theta_\text{r}}$. This coarse-to-fine pipeline allows WarpingNet to learn a fine-grained image sampling grid $\Omega$ to warp the projector input image $x$ to the camera's canonical frontal view by $\phi(x, \Omega)$, where $ \phi(\cdot; \cdot) $ is a differentiable image interpolator \cite{jaderberg2015spatial} denoted as $ \otimes $. Then, we use the input projector direct light mask $I_\text{m}$ to exclude occluded pixels by $\phi(x, \Omega)\odot I_\text{m}$, where $\odot$ is element-wise multiplication. Afterwards, this warped projector image is further used to compute an intermediate rough shading image $\phi(x, \Omega)\odot I_\text{m}\odot I_s$ to enforce the occlusion constraint. 
		\textbf{ShadingNet} has a two-branch encoder-decoder structure to capture complex photometric transformations.
		In particular, it concatenates $I_s$ and $\phi(x, \Omega)\odot I_\text{m}\odot I_s$ and feeds them to the middle encoder branch. Similarly, $\phi(x, \Omega)\odot I_\text{m}$ is fed to the backbone encoder branch. The skip connections between the two branches model photometric interactions between the three inputs at different levels. In addition, we pass $I_s$ to the output layer through three convolutional layers.
		Finally, the feature maps are fused into one inferred camera-captured scene (under superimposed projection) $\hat{I}_x$ by the backbone decoder.
		}\label{fig:flowchart}
	\end{center}
\end{figure*}

This optimization problem involves the real project-and-capture process $ \pi $, and it has no analytical gradient. Theoretically, we can compute numerical gradient instead, but it is extremely inefficient, \eg, for a $256\times256$ projector resolution, $256\times256\times3$ real project-and-capture processes are required to compute the Jacobian matrix for a single gradient descent step. 
To avoid gradient computation and reduce project-and-capture processes, Nichols and Jasper \cite{nichols2018ProjectingTL} include $ \pi $ in a gradient-free optimization (\eg, differential evolution) and only perturb one projector pixel. However, even for a low resolution image (\eg, $32\times32$), hundreds of real project-and-capture processes are required for a single targeted attack, let alone for higher resolutions. Moreover, because only one-pixel perturbation is allowed, this method also suffers from low attack success rates when image resolution increases.

Another intuitive solution is to digitally attack the camera-captured scene image under normal light first, \ie, $I_{x_0}+\delta$ (\autoref{eq:digital_tar}), then use a projector compensation method, \eg, CompenNet++ \cite{huang2019compennet++}, to find its corresponding projector input image by: $x' = \pi^{\dagger}(I_{x_0}+\delta)$, where $\pi^{\dagger}:I_x \mapsto x$ (named CompenNet++) is the pseudo-inverse of $\pi$. However, digital attacks are unaware of the physical constraints of the projector-camera system (\eg, dynamic ranges and occlusions), thus the generated digital adversarial image $I_{x_0}+\delta$ may contain physically implausible perturbations. Therefore, even if $\pi^{\dagger}$ is a perfect approximation of $\pi$'s inverse, the real camera-captured scene under superimposed projection may not match the generated digital version. Moreover, CompenNet++ cannot address occlusions and those regions may become blurry after compensation.

In this paper, we propose a more practical and accurate solution by first approximating the real project-and-capture process $ \pi $ with a deep neural network, named PCNet $ \hat{\pi}_\theta $ parameterized by $\theta$. Then, we substitute the real project-and-capture process $ \pi $ with PCNet $ \hat{\pi}$ in \autoref{eq:real_pc_obj}. Finally, fixing the weights of the classifier $ f $ and PCNet $ \hat{\pi} $, the projector adversarial image $ x' $ can be solved by optimizing \autoref{eq:real_pc_obj} using gradient descent. Our approach brings three advantages: (a) because PCNet $ \hat{\pi} $ is differentiable, we can use analytical gradient to improve adversarial attack optimization efficiency; (b) Compared with two-step methods, \eg, digital attack with projector compensation, PCNet can model physical constraints of the projector-camera system, thus it can produce more robust and stealthy adversarial attacks; (c) Because PCNet can be trained offline, it requires only one online project-and-capture process for stealthy projector-based attacks.

\subsection{PCNet $ \hat{\pi} $}
\noindent\textbf{Formulation}.
In \autoref{eq:real_pc_obj}, the real project-and-capture process $\pi$ takes three inputs, \ie, a projector input image $x$, the environment light $l$ and the physical scene $s$. For each setup, $l$ and $s$ remain static,  and only the projector input image $x$ is varied, thus we can approximate $l$ and $s$ with a camera-captured image $I_s = I_{x_0} = \pi(x_0, l, s)$. In practice, the camera may suffer from large sensor noise under low light, thus we set $x_0$ to a plain gray image to provide some illumination, \ie, $x_0= [128, 128, 128]^{256\times 256}$. Another practical issue is occlusion, which may jeopardize PCNet training and adversarial attack if not properly modeled. Thus, we explicitly extract a projector direct light mask $I_\text{m}$ using the method in \cite{nayar2006fast}. 
Then, the camera-captured scene under superimposed projection can be approximated by:
\begin{equation}\label{eq:PCNet}
	\hat{I}_x = \hat{\pi}(x, I_s, I_\text{m})
\end{equation}
Apparently $\hat{\pi}$ implicitly encodes both geometric and photometric transformations between the projector input and camera-captured images, and may be learned using a general image-to-image translation network. However, previous work (\eg, \cite{huang2019compennet++}) shows that explicitly disentangling geometry and photometry significantly improves network convergence, especially for limited training data and time.

\vspace{1.5mm}\noindent\textbf{Network design}.
As shown in \autoref{fig:flowchart}, PCNet consists of two subnets: \textbf{WarpingNet} (for geometry) and \textbf{ShadingNet} (for photometry), and this architecture is inspired by CompenNet++ \cite{huang2019compennet++}, which uses a CNN for projector compensation by learning the \textit{backward} mapping $\pi^{\dagger}: I_x \mapsto x$. By contrast, our PCNet learns the \textit{forward} mapping (\ie, $\pi:x\mapsto I_x  $) from a projector input image $x$ to the camera-captured scene under superimposed projection. In addition, CompenNet++ is designed for smooth surfaces, and it assumes no occlusions in camera-captured images, thus it may not work well if directly applied to stealthy projector-based attacks where occlusions exist. As shown in our experiments, CompenNet++ produces strong artifacts on our setups (\autoref{fig:exp_tar}), while our PCNet addresses this issue by inputting an additional 
projector direct light mask $I_\text{m}$ to exclude occluded pixels. Moreover, we compute a rough shading image $\phi(x, \Omega)\odot I_\text{m}\odot I_s$ as an additional input for ShadingNet, and it brings improved performance compared with CompenNet++'s photometry part (\ie, CompenNet).

Finally, for each scene $ s $ under lighting $ l $, given a camera-captured scene image $I_s$, a projector direct light mask $I_\text{m}$ and projected and captured image pairs $\{(x_i, I_{x_i})\}_{i=1}^{M}$, PCNet parameters $\theta$ (\ie, pink blocks in \autoref{fig:flowchart}) can be trained using image reconstruction loss $\mathcal{L}$ (\eg,  pixel-wise $L_1$+SSIM loss \cite{zhao2017loss}) below:
\begin{equation}\label{eq:procam_loss}
	\theta = \argmin_{\theta'}\sum_i\mathcal{L}\big(\hat{I}_{x_i} = \hat{\pi}_{\theta'}(x_i, I_s, I_\text{m}), ~I_{x_i}\big)
\end{equation}

We implement PCNet using PyTorch \cite{paszke2017automatic} and optimize it using Adam optimizer \cite{kinga2015method} for 2,000 iterations with a batch size of 24, and it takes about 6.5 minutes to finish training on three Nvidia GeForce 1080Ti GPUs.

\setlength{\algomargin}{0em} 
\begin{algorithm}[t]
\SetAlCapHSkip{0em} 
\SetAlgoLined 
\SetKwComment{Comment}{$\triangleright$\ }{} 
\SetSideCommentLeft
\SetAlgoLongEnd 
\DontPrintSemicolon 
\SetKwInput{Input}{Input}
\SetKwInOut{Output}{Output}
\SetKw{KwOr}{or}

\caption{SPAA: Stealthy Projector-based Adversarial Attack.}\label{alg:alternating}
\Input{\;
$x_0$: projector plain gray image\;
$I_s$: camera-captured scene under $x_0$ projection\;
$I_\text{m}$: projector direct light mask\;
$t$: target class\;
$K$: number of iterations\;
$p_\text{thr}$: threshold for adversarial confidence\;
$d_\text{thr}$: threshold for $L_2$ perturbation size\;
$\beta_1$: step size in minimizing adversarial loss\;
$\beta_2$: step size in minimizing stealthiness loss\;
}
\Output{
$x'$: projector adversarial image}
\vspace{1.5mm}Initialize $x'_0 \gets x_0$\;
\For{$k\gets1$ \KwTo $K$}{
	$\hat{I}_{x'} \gets \hat{\pi}(x'_{k-1}, I_s, I_\text{m})$\;
	$d \gets \|\hat{I}_{x'}-I_s\|_2$\;
  \eIf{$f_t(\hat{I}_{x'}) < p_\text{thr}$ \KwOr $d<d_\text{thr}$}
  {
    $g_1\gets\alpha\nabla_{x'} f_t(\hat{I}_{x'})$ \hspace{3em}\hfill// minimize adversarial loss\; 
    $x'_k\gets x'_{k-1}+\beta_1*\frac{g_1}{\norm{g_1}_2}$\;
  }
  {  
     $g_2\gets-\nabla_{x'} d $  \hspace{5em}\hfill// minimize stealthiness loss\; 
    $x'_k\gets x'_{k-1}+\beta_2*\frac{g_2}{\norm{g_2}_2}$\;
  }
  $x'_k\gets \text{clip}(x'_k, 0, 1)$\; 
 }
\KwRet $x'\gets x'_k$ that is adversarial and has smallest $d$
\end{algorithm}

\subsection{Stealthy projector-based adversarial attack}
Once PCNet $ \hat{\pi} $ is trained, we replace the real project-and-capture process $ \pi $ in \autoref{eq:real_pc_obj} by $ \hat{\pi} $ using \autoref{eq:PCNet}, then stealthy projector-based adversarial attacks are to find an image $ x' $ such that
\begin{align}\label{eq:real_pc_obj_raw}
	&   \argmax_i f_i\left(I_{x'} = \hat{\pi}(x', I_s, I_\text{m})\right)
		\begin{cases*}
		 = t, \quad\quad &\text{targeted}\\
		\neq t_\text{true} \quad\quad & \text{untargeted}
		\end{cases*} \nonumber\\
	& \text{subject to} \quad \mathcal{D}\left(I_{x'}, \ I_s\right) < \epsilon
\end{align}
Here, we choose $L_2$ norm as our image distance/stealthiness metric $\mathcal{D}$, results on other image distance metrics such as $\Delta E$ and $\Delta E+L_2$ can be found in the supplementary. 
Then, we propose to solve \autoref{eq:real_pc_obj_raw} by minimizing the following loss function with gradient descent:
\begin{equation}\label{eq:attack_loss}
	x' = \argmin_{x'}\underbrace{\alpha f_t(I_{x'})}_{\text{adversarial loss}} + 
		\underbrace{\norm{I_{x'} - I_s}_2}_{\text{stealthiness loss}}
\end{equation}
where $\alpha=-1$ for targeted attacks and $\alpha=1$ for untargeted attacks. 

To get higher attack success rates while remaining stealthy, we develop an optimization algorithm (\autoref{alg:alternating}) that alternates between the adversarial loss and stealthiness loss in \autoref{eq:attack_loss}. Note that our method is inspired by digital attack algorithms PerC-AL \cite{zhao2020towards} and DDN \cite{rony2019decoupling} with the following differences: (a) PerC-AL  and DDN are digital attacks while our algorithm is designed for projector-based attacks by including a deep neural network approximated project-and-capture process $\hat{\pi}$; (b) We add two hyperparameters, perturbation size threshold $d_\text{thr}$ and adversarial confidence threshold $p_\text{thr}$ to improve transferability from $\hat{\pi}$ to $\pi$. It is worth noting that we have tried simply optimizing the weighted sum of adversarial and stealthiness losses, and it led to an inferior performance compared with the alternating algorithm.

For \autoref{alg:alternating}, we initialize $ x' $ with a projector plain gray image $ x_0 $ and run optimization for $K=50$ iterations. After experiments on different settings, we set the step sizes to $\beta_1 = 2, \beta_2=1$. The adversarial confidence threshold is set to $p_\text{thr}=0.9$ and the perturbation size threshold $d_\text{thr}$ is varied from 5 to 11 (\autoref{sec:additional_results}). Note that \autoref{alg:alternating} is highly parallelizable and multi-classifier and multi-targeted attacks can simultaneously run in batch mode.

\begin{table*}[!t]
\begin{center}
\caption{Quantitative comparison of projector-based adversarial attacks on Inception v3 \cite{szegedy2016rethinking}, ResNet-18 \cite{he2016deep} and VGG-16 \cite{simonyan2014very}. Results are averaged on 13 setups. The top section shows our SPAA results with different thresholds for $L_2$ perturbation size $d_\text{thr}$ as mentioned in \autoref{alg:alternating}. The bottom section shows two baselines \ie, PerC-AL+CompenNet++ \cite{zhao2020towards, huang2019compennet++} and One-pixel DE \cite{nichols2018ProjectingTL}. The 4\textsuperscript{th} to 6\textsuperscript{th} columns are targeted (T) and untargeted (U) attack success rates, and the last four columns are stealthiness metrics. Please see \textbf{supplementary} for more results.}\label{tab:compare}
\begin{tabular}{@{}ccrccccccc@{}}
	\toprule[0.5mm]
	                                                                          &                                              & \textbf{Classifier} & \textbf{T. top-1 (\%)} & \textbf{T. top-5 (\%)} & \textbf{U. top-1 (\%)} & $L_2\downarrow$ & $L_\infty\downarrow$ & $\Delta E$$ \downarrow $ & \textbf{SSIM}$ \uparrow $ \\ \midrule[0.3mm]
	                                                                          &              $d_\text{thr} =5$               &        Inception v3 &         41.54          &         67.69          &         84.62          &      6.273      &        5.101         &          2.588           &           0.937           \\
	                                                                          &                                              &           ResNet-18 &         73.08          &         90.00          &         100.00         &      6.304      &        5.158         &          2.701           &           0.940           \\
	                                                                          &                                              &              VGG-16 &         69.23          &         83.85          &         100.00         &      6.629      &        5.428         &          2.824           &           0.934           \\
	\cmidrule[0.3mm]{2-10}        \multirow{8}{*}{\rotatebox{90}{Our method}} &              $d_\text{thr} =7$               &        Inception v3 &         67.69          &         84.62          &         100.00         &      7.603      &        6.199         &          3.135           &           0.904           \\
	                                                                          &                                              &           ResNet-18 &         92.31          &         94.62          &         100.00         &      7.786      &        6.396         &          3.349           &           0.907           \\
	                                                                          &                                              &              VGG-16 &         83.08          &         97.69          &         100.00         &      8.117      &        6.668         &          3.435           &           0.899           \\
	                         \cmidrule[0.3mm]{2-10}                           &              $d_\text{thr} =9$               &        Inception v3 &         76.15          &         90.00          &         100.00         &      9.336      &        7.620         &          3.766           &           0.872           \\
	                                                                          &                                              &           ResNet-18 &         95.38          &         98.46          &         100.00         &      9.640      &        7.923         &          4.066           &           0.874           \\
	                                                                          &                                              &              VGG-16 &         90.00          &         99.23          &         100.00         &      9.978      &        8.211         &          4.156           &           0.864           \\
	                         \cmidrule[0.3mm]{2-10}                           &              $d_\text{thr} =11$              &        Inception v3 &         76.92          &         92.31          &         100.00         &     11.190      &        9.156         &          4.386           &           0.843           \\
	                                                                          &                                              &           ResNet-18 &         97.69          &         100.00         &         100.00         &     11.605      &        9.545         &          4.785           &           0.846           \\
	                                                                          &                                              &              VGG-16 &         94.62          &         99.23          &         100.00         &     11.750      &        9.671         &          4.784           &           0.835           \\ \midrule[0.5mm]
	               \multirow{6}{*}{\rotatebox{90}{Baselines}}                 &             PerC-AL+CompenNet++              &        Inception v3 &         20.00          &         42.31          &         84.62          &      7.430      &        6.006         &          2.690           &           0.949           \\
	                                                                          & \cite{zhao2020towards,huang2019compennet++}  &           ResNet-18 &         40.77          &         52.31          &         100.00         &      7.713      &        6.249         &          2.823           &           0.943           \\
	                                                                          &                                              &              VGG-16 &         33.85          &         49.23          &         100.00         &      7.526      &        6.099         &          2.753           &           0.946           \\
	                         \cmidrule[0.3mm]{2-10}                           & One-pixel DE  \cite{nichols2018ProjectingTL} &        Inception v3 &          0.00          &          1.54          &         15.38          &      8.388      &        6.550         &          2.460           &           0.973           \\
	                                                                          &                                              &           ResNet-18 &          0.00          &          0.00          &          7.69          &      8.034      &        6.276         &          2.401           &           0.976           \\
	                                                                          &                                              &              VGG-16 &          0.00          &          1.54          &         23.08          &      8.233      &        6.410         &          2.473           &           0.975           \\ \bottomrule[0.5mm]
\end{tabular}
\end{center}
\end{table*}

\section{Experimental Evaluations}\label{sec:experiments}
\subsection{System configurations}
Our setup consists of a Canon EOS 6D camera and a ViewSonic PA503S DLP projector, as shown in \autoref{fig:teaser}. Their resolutions are set to $320\times240$ and $800\times600$, respectively. The projector input image resolution is set to $256\times256$. The distance between the projector-camera pair and the target object is around 1.5 meters.

Note that PCNet is trained/tested individually for each setup. We capture 13 different setups with various objects (see supplementary). For each setup, we first capture a scene image $I_s$ and two shifted checkerboard patterns to extract the scene direct illumination component using the method in~\cite{nayar2006fast}, and obtain the projector direct light mask $I_\text{m}$ by thresholding the direct illumination component. Then, we capture $M=500$ sampling image pairs $\{(x_i, I_{x_i})\}_{i=1}^{M}$ (took 3 minutes) for training PCNet $\hat{\pi}$. 
Afterwards, for each setup we apply \autoref{alg:alternating} to ten projector-based targeted attacks and one untargeted attack on three classifiers \ie, ResNet-18 \cite{he2016deep}, VGG-16 \cite{simonyan2014very} and Inception v3 \cite{szegedy2016rethinking}. In total, it takes 34 seconds to generate the adversarial projection patterns and another 17 seconds to project and capture all of them.

\begin{figure*}[!t]
	\begin{center}
		\includegraphics[width=1\linewidth]{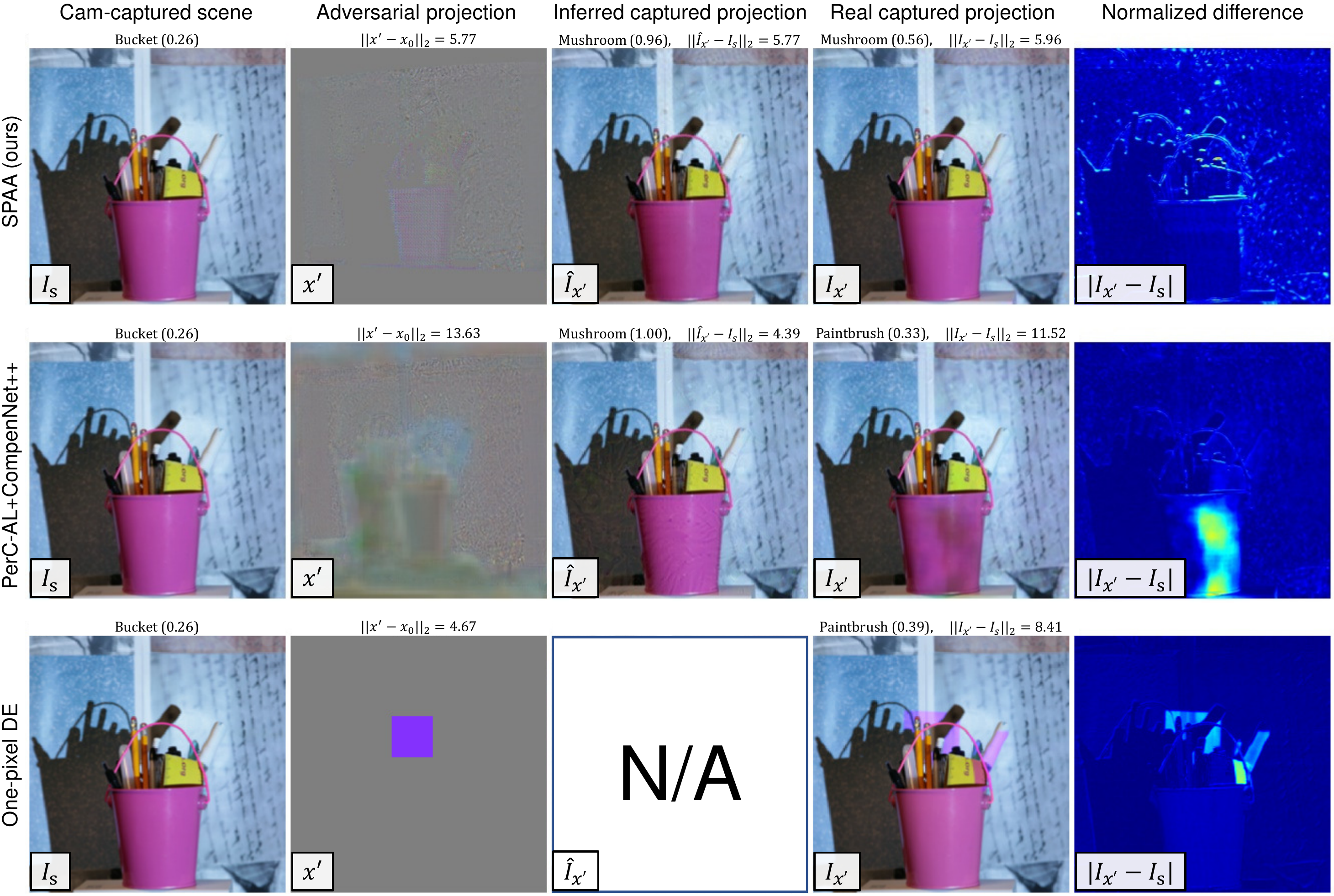}
		\caption{\textbf{Targeted projector-based adversarial attack on VGG-16}. The goal is to use adversarial projections to cause VGG-16 to misclassify the camera-captured scene as \textbf{mushroom}. The 1\textsuperscript{st} to the 3\textsuperscript{rd} rows are our SPAA, PerC-AL + CompenNet++ \cite{zhao2020towards,huang2019compennet++} and One-pixel DE \cite{nichols2018ProjectingTL}, respectively. The 1\textsuperscript{st} column shows the camera-captured scene under plain gray illumination. The 2\textsuperscript{nd} column shows inferred projector input adversarial patterns. The 3\textsuperscript{rd} column plots model inferred camera-captured images. The 4\textsuperscript{th} column presents real captured scene under adversarial projection \ie, the 2\textsuperscript{nd} column projected onto the 1\textsuperscript{st} column. The last column provides normalized differences between the 4\textsuperscript{th} and 1\textsuperscript{st} columns. On the top of each camera-captured image, we show the classifier's predicted labels and probabilities. For the 2\textsuperscript{nd} to 4\textsuperscript{th} columns, we also show the $L_2$ norm of  perturbations. Note that for One-pixel DE, the 3\textsuperscript{rd} column is blank because it is an online method and no inference is available. Note that both baselines fail in this \emph{targeted} attack. Please see \textbf{supplementary} for more results. }\label{fig:exp_tar}
	\end{center}
\end{figure*}

\begin{figure*}[!t]
	\begin{center}
		\includegraphics[width=1\linewidth]{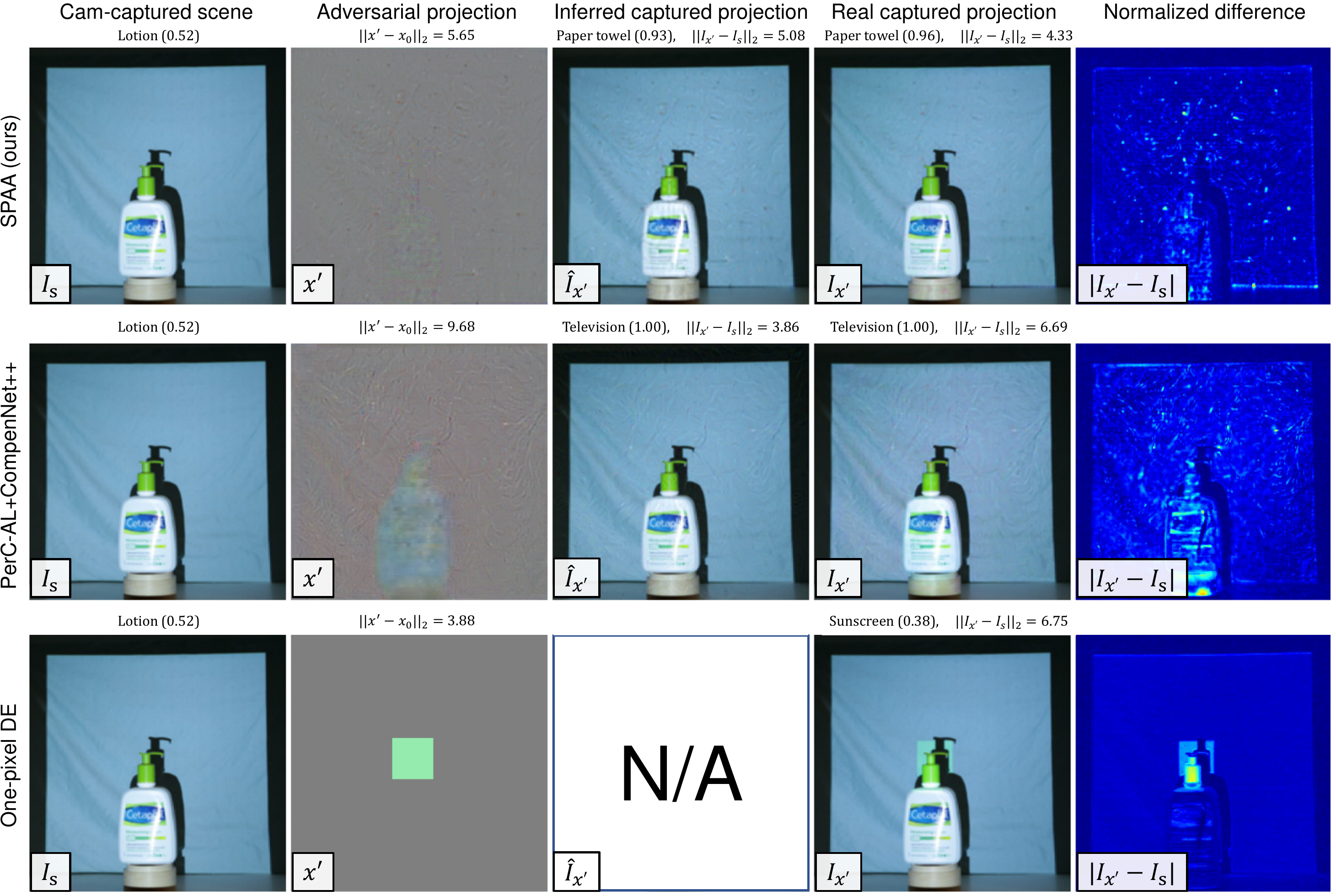}
		\caption{\textbf{Untargeted projector-based adversarial attack on Inception v3}. The goal is to use adversarial projections to cause Inception v3 to misclassify the camera-captured scene as any label other than \textbf{lotion}. The 1\textsuperscript{st} to the 3\textsuperscript{rd} rows are our SPAA, PerC-AL + CompenNet++ \cite{zhao2020towards,huang2019compennet++} and One-pixel DE \cite{nichols2018ProjectingTL}. On the top of each camera-captured image, we show the classifier's predicted labels and probabilities. For the 2\textsuperscript{nd} to 4\textsuperscript{th} columns, we also show the $L_2$ norm of  perturbations. Note that for One-pixel DE, the 3\textsuperscript{rd} column is blank because it is an online method and no inference is available. See \textbf{supplementary} for more results. }\label{fig:exp_untar}
	\end{center}
\end{figure*}

\subsection{Evaluation benchmark}
We evaluate stealthy projector-based attack methods by targeted and untargeted attack success rates and stealthiness measured by similarities between the camera-captured scene $I_s$ and the camera-captured scene under adversarial projection $I_{x'}$ using $L_2$ norm, $L_\infty$ norm, perceptual color distance $\Delta E$ \cite{luo2001development} and SSIM \cite{wang2004image}. 

We first compare with the gradient-free differential evolution (DE)-based baseline \cite{nichols2018ProjectingTL}, named \textit{One-pixel DE}, which only alters one projector pixel. Originally, it was designed for attacking classifiers trained on $32\times32$ CIFAR-10 \cite{krizhevsky2009learning} images, with both the projector and camera resolutions set to $32\times32$ as well. However, as shown in the last three rows of \autoref{tab:compare}, the top-1 targeted attack success rates are 0, meaning that in our higher resolution setups, this method failed to fool the three classifiers (ResNet-18 \cite{he2016deep}, VGG-16 \cite{simonyan2014very} and Inception v3 \cite{szegedy2016rethinking}) trained on ImageNet \cite{deng2009imagenet}. To increase its attack success rates, we increase the original perturbed projector pixel size from $1\times1$ to $41\times41$, and then we see a few successful untargeted attacks. In terms of efficiency, we use the same DE parameters as \cite{nichols2018ProjectingTL}, and it takes one minute to attack a single image and 33 minutes to attack three classifiers in total, while our method only takes 10 minutes including PCNet training, adversarial attack and real project-and-capture. Note that our method can simultaneously attack multiple classifiers and targets while \textit{One-pixel DE} involves a non-parallelizable real project-and-capture process, and this advantage may become more significant when the numbers of adversarial targets and classifiers increase.

We then compare with a two-step baseline that first performs digital attacks on the camera-captured image by $\hat{I}_x = I_s + \delta$. For this step, we adapt the state-of-the-art PerC-AL \cite{zhao2020towards} to our projector-based attack problem. The original PerC-AL assumes a just sufficient adversarial effect, \ie, the generated digital adversarial examples just successfully fool the classifiers without pursuing a higher adversarial confidence. However, in our task, these examples failed to fool the classifiers after real project-and-capture processes, due to the complex physical environment and the image capture process of projector-based attacks. Thus, similar to our SPAA, we add an adversarial confidence threshold $p_\text{thr}$ to PerC-AL's optimization to allow this algorithm to pursue a more robust adversarial attack, \ie, a digital adversarial example is adversarial only when its probability is greater than $p_\text{ptr}$. Then we use CompenNet++ \cite{huang2019compennet++} to find the corresponding projector adversarial image $x' = \pi^{\dagger}(\hat{I}_x, I_s)$. In practice, CompenNet++ is trained using the same sampling image pairs as PCNet, but with the network input and output swapped. Moreover, unlike PCNet, CompenNet++ does not use occlusion mask $I_m$ or compute a rough shading image. We name this method \textit{PerC-AL + CompenNet++}. Note that we do not compare with \cite{zhou2018invisible, shen2019vla} because they are specifically designed for faces only.

\vspace{1mm}\noindent\textbf{Quantitative comparisons}. As shown in \autoref{tab:compare}, the proposed SPAA significantly outperforms \textit{One-pixel DE} \cite{nichols2018ProjectingTL} and the two-step \textit{PerC-AL + CompenNet++} \cite{zhao2020towards,huang2019compennet++} by having higher attack success rates (the 4\textsuperscript{th} to 6\textsuperscript{th} columns of \autoref{tab:compare}) and stealthiness ($L_2$ and $L_\infty$). 
Note that \textit{One-pixel DE} has very low targeted attack success rates, because it only perturbs a $41\times41$ projector image block, and such camera-captured images have strong square patterns (see the 3\textsuperscript{rd} row of \autoref{fig:exp_tar}) that are clearly far from the adversarial target image distributions, they are also less stealthy. In our experiments, we find \textit{One-pixel DE} can reduce the confidence of the true label,  but it can rarely increase the probability of a specific adversarial target, because te projected color square is too simple. Moreover, digital targeted attacks on classifiers trained on ImageNet ($224\times224$, 1,000 classes) are already much harder than those trained on CIFAR-10 ($32\times32$, 10 classes), due to higher image resolutions and $100$ times more classes, let alone applying it to the more challenging stealthy projector-based attacks. 
By contrast, our SPAA and \textit{PerC-AL + CompenNet++} have higher success rates and stealthiness than \textit{One-pixel DE}. These results are also shown in qualitative comparisons below.

\noindent\textbf{Qualitative comparisons}. Exemplar projector-based \textit{targeted} and \textit{untargeted} adversarial attack results are shown in \autoref{fig:exp_tar} and \autoref{fig:exp_untar}, respectively. 
In \autoref{fig:exp_tar}, clearly our method can achieve successful attacks while remaining stealthy. \textit{PerC-AL + CompenNet++} failed this targeted attack, and we see two particular problems: \textbf{(1)} it produces a blurry bucket-like projection pattern (2\textsuperscript{nd} row, 2\textsuperscript{nd} column), because CompenNet++ cannot learn compensation well under occlusions. Thus, when the adversarial pattern is projected to the scene, we see large dark artifacts on the bucket (2\textsuperscript{nd} row, 4\textsuperscript{th}-5\textsuperscript{th} columns). By contrast, our SPAA addresses occlusions by computing a projector direct light mask, then explicitly generates a rough shading image to enforce the occlusion constraint. Clearly, our generated adversarial projections (1\textsuperscript{st} row, 2\textsuperscript{nd} column) show much weaker artifacts.
\textbf{(2)} We also see strong adversarial patterns in the bucket shadow (2\textsuperscript{nd} row, 3\textsuperscript{rd} column), however, the projector is unable to project to this occluded region. This is caused by the first step that performs a digital attack by $\hat{I}_x = I_s + \delta$. Without any prior knowledge about the real project-and-capture process, this step may generate physically implausible adversarial patterns like this. 
By contrast, our SPAA uses an end-to-end differentiable formulation, with which we include a neural network approximated project-and-capture process, \ie, PCNet in the projector-based attack optimization. Then, physical constraints are explicitly applied, such that the generated adversarial pattern is physically plausible. Thus, we do not see undesired adversarial patterns in the bucket shadow of the 1\textsuperscript{st} row, 3\textsuperscript{rd} column.

For untargeted attacks, as shown in the 4\textsuperscript{th} column of \autoref{fig:exp_untar}, all three methods successfully fooled Inception v3 \cite{szegedy2016rethinking}, as the classifier predicted labels are NOT \textbf{lotion}. In addition, compared with the two baselines, our method has the smallest perturbation size ($L_2$ norm is 4.33), and the projected adversarial image (the 2\textsuperscript{nd} column) and camera-captured adversarial projection (the 4\textsuperscript{th} column) are also stealthier. 
More untargeted attack results can be found in the \textbf{supplementary} Figures 14-26, where \textit{One-pixel DE} \cite{nichols2018ProjectingTL} shows successful untargeted attacks in Figures 14 and 16. For other scenes, although \textit{One-pixel DE} \cite{nichols2018ProjectingTL} failed untargeted attacks, it decreases the classifiers' confidence of the true labels.

\begin{table}[!b]
	\begin{center}
	\caption{Quantitative comparisons between \textbf{PCNet} and PCNet without the direct light mask and rough shading image (\textbf{PCNet w/o mask and rough}). The image similarity metrics below are calculated between the real camera-captured scene under adversarial projection $I_x$ (GT) and the model inferred camera-captured scene under adversarial projection $\hat{I}_x$. Results are averaged on 13 setups.}\label{sec:pcnet}
	\begin{tabular}{@{}lcccc@{}}
	\toprule[0.5mm]
	\textbf{Model name}      & $L_2\downarrow$ & $L_\infty\downarrow$ & $\Delta E\downarrow $ & \textbf{SSIM}$\uparrow $ \\ \midrule[0.3mm]
	PCNet                    & 10.461                                  & 8.408                                        & 3.066                                         & 0.947                                        \\
	PCNet w/o mask and rough & 11.952                                  & 9.567                                        & 3.385                                         & 0.932                                        \\ \bottomrule[0.5mm]
	\end{tabular}
	\end{center}
\end{table}

\subsection{Perturbation size threshold and PCNet components}\label{sec:additional_results}
In this section, we study the proposed SPAA's success rates with different perturbation size thresholds ($ d_\text{thr} $) and the effectiveness of PCNet's direct light mask and rough shading image. For comparisons on different stealthiness loss functions, we refer the readers to the supplementary.

\noindent\textbf{Perturbation size threshold} $d_\text{thr}$ is the minimum perturbations of the PCNet $ \hat{\pi} $ inferred camera-captured scene under adversarial projection. As shown in \autoref{alg:alternating}, a higher $d_\text{thr}$ can lead to a stronger adversary and higher projector-based attack success rates. In \autoref{tab:compare}, we show different $d_\text{thr}$ ranging from 5 to 11. Clearly, attack success rates and real camera-captured perturbation sizes (\ie, $L_2$, $L_\infty$,  $\Delta E$ and SSIM) increase as $d_\text{thr}$ increases. Thus, it controls the trade-off between projector-based attack success rates and stealthiness.

\vspace{1mm}\noindent\textbf{PCNet direct light mask and rough shading image.} For each setup, we project and capture 200 colorful and textured images $x$, then we compare the similarities between the real camera-captured scene under adversarial projection $I_x$ and PCNet inferred camera-captured scene under adversarial projection $\hat{I}_x$ using $L_2$ norm, $L_\infty$ norm, $\Delta E$ and SSIM. The results are shown in \autoref{sec:pcnet} and PCNet outperforms the degraded version that is without direct light mask and rough shading image, demonstrating that we need to model the essential factors, \ie, direct light mask and rough shading image for better project-and-capture approximation.


\section{Conclusion}\label{sec:conclusion}
In this paper, for the first time, we formulate stealthy projector-based adversarial attack as an end-to-end differentiable process, and propose a solution named SPAA (Stealthy Projector-based Adversarial Attack). In SPAA, we approximate the real project-and-capture process using a deep neural network named PCNet (Project-And-Capture Network), which not only allows the gradients to backpropagate to the projector input adversarial pattern, but also provides additional physical constraints for adversarial attack optimization, such that the generated adversarial projection is physically plausible. In addition, we propose an algorithm to alternate between the adversarial loss and stealthiness loss using minimum perturbation and adversarial confidence thresholds. In our thorough experiments, SPAA significantly outperforms other methods by significantly higher attack success rates and stealthiness, for both targeted and untargeted attacks.

\noindent\textbf{Limitations and future work.} Although our PCNet can better model the project-and-capture process than CompenNet++ \cite{huang2019compennet++}, it is not perfect, and we can see some discrepancies between the simulated and the real attacks in \autoref{fig:teaser} (d) and (e). In future work, we can improve PCNet by incorporating physically based rendering domain knowledge in network design. Another limitation of our SPAA is its sensitivity to environment light, and  improving its robustness under different light conditions is also an interesting direction to explore in the future.

\noindent\textbf{Acknowledgements.} We thank the anonymous reviewers for valuable and inspiring comments and suggestions.
\bibliographystyle{abbrv-doi}

\bibliography{ref}
\end{document}